\documentclass{article}
\usepackage[preprint]{spconf,amsmath,graphicx}
\usepackage{amsfonts}
\usepackage{booktabs}

\title{Federated Learning with Instance-Dependent Noisy Label}
\name{Lei Wang, Jieming Bian, Jie Xu}
\address{Electrical and Computer Engineering, University of Miami, 
Coral Gables, FL 33146, USA}
\usepackage{tikz}

\newcommand\copyrighttext{%
  \footnotesize \textcopyright 2024 IEEE. Personal use of this material is permitted. Permission from IEEE must be obtained for all other uses, including reprinting/republishing this material for advertising or promotional purposes, collecting new collected works for resale or redistribution to servers or lists, or reuse of any copyrighted component of this work in other works.}

\copyrightnotice{%
\begin{tikzpicture}[remember picture,overlay]
\node[anchor=south,yshift=10pt] at (current page.south) {\fbox{\parbox{\dimexpr0.75\textwidth-\fboxsep-\fboxrule\relax}{\copyrighttext}}};
\end{tikzpicture}%
}

\begin{document}
\ninept
\maketitle
%
%

\begin{abstract}
Federated learning (FL) with noisy labels poses a significant challenge. Existing methods designed for handling noisy labels in centralized learning tend to lose their effectiveness in the FL setting, mainly due to the small dataset size and the heterogeneity of client data. While some attempts have been made to tackle FL with noisy labels, they primarily focused on scenarios involving class-conditional noise. In this paper, we study the more challenging and practical issue of instance-dependent noise (IDN) in FL. We introduce a novel algorithm called FedBeat (Federated Learning with Bayesian Ensemble-Assisted Transition Matrix Estimation). FedBeat aims to build a global statistically consistent classifier using the IDN transition matrix (IDNTM), which encompasses three synergistic steps: (1) A federated data extraction step that constructs a weak global model and extracts high-confidence data using a Bayesian model ensemble method. (2) A federated transition matrix estimation step in which clients collaboratively train an IDNTM estimation network based on the extracted data. (3) A federated classifier correction step that enhances the global model's performance by training it using a loss function tailored for noisy labels, leveraging the IDNTM. Experiments conducted on CIFAR-10 and SVHN verify that the proposed method significantly outperforms state-of-the-art methods.

\end{abstract}


\section{Introduction}

Advancements in mobile edge devices have led to the collection of massive datasets, sparking significant interest in utilizing this data for training deep learning models. However, leveraging such mobile device data for deep learning poses a complex challenge due to privacy concerns. To address this issue, Federated Learning (FL) has emerged as a promising research area that enables clients with decentralized data to collaborate and train a shared model under the guidance of a central server. Introduced by \cite{mcmahan2017communication}, FedAvg has been instrumental in reducing server-client communication while preserving data privacy through local stochastic gradient descent. Various researchers have extended the FedAvg framework to other real-world scenarios. For instance, some studies \cite{zhao2018federated, sattler2019robust, li2020federated} address the challenge of heterogeneous client data, while others \cite{cho2020client, karimireddy2020scaffold, wang2022unified} focus on mitigating the effects of partial client participation.



\begin{figure}
  \centering
  \includegraphics[width=0.45\textwidth]{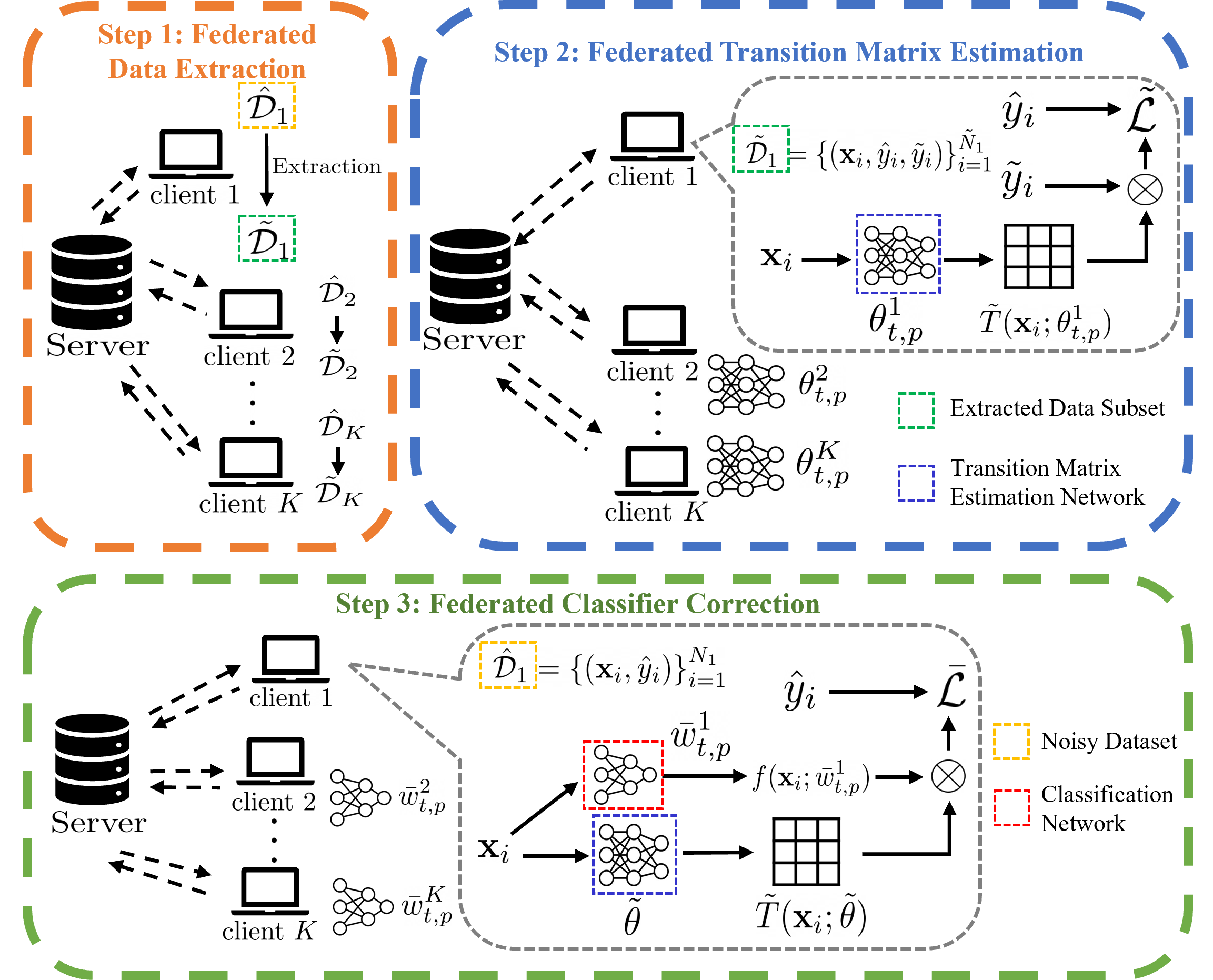}
  \vspace{-5pt}
  \caption{An overview of FedBeat, presented in a three-step structure.}
  \label{overview}
  \vspace{-15pt}
\end{figure}

While significant progress has been made in adapting FL to realistic settings, the issue of label quality associated with data collected by edge devices has often been overlooked. Unlike traditional centralized learning, where expert-provided labels are available for each data sample, ensuring label quality in FL is considerably more challenging. Many users lack expertise, making it unreasonable to expect all data samples to have ground truth labels. Consequently, addressing the problem of noisy labels in FL is of paramount importance.

In centralized settings, extensive research has been conducted on learning with noisy data. Some researchers \cite{luengo2018cnc, wei2020combating, li2019learning} have proposed noise-cleaning approaches that aim to detect noisy samples and train with clean ones. Others \cite{cheng2022instance, yang2022estimating, xia2020part, chen2021beyond} have estimated transition matrices to build statistically consistent classifiers, reducing the impact of noisy labels by inferring clean distributions based on the transition matrix and noisy class-posterior probabilities. However, unlike the sufficient dataset size in centralized learning, the dataset stored on each client in FL is relatively small, making it impossible for individual clients to train high-performance models without collaboration. Moreover, the heterogeneity of clients' data distributions complicates the collaboration process. Only a few studies \cite{xu2022fedcorr, yang2022robust, tsouvalas2022federated} have attempted to address FL with noisy labels. However, these methods primarily focus on class-conditional noise (CCN), where each instance from one class has a fixed probability of being assigned to another. In reality, noisy labels are often instance-based, indicating that a data sample is more likely to be mislabeled due to its content rather than the class label it belongs to. This form of noise is commonly referred to as instance-dependent noise (IDN).

In this paper, we focus on addressing IDN in the context of FL, as it is more relevant for handling real-world noisy scenarios \cite{cheng2022instance} and a previously unexplored problem. Our key idea relies on constructing an instance-dependent noise transition matrix (IDNTM) estimation network to establish the correspondence between the clean label and the noisy label, which subsequently enables the training of a statistically consistent classifier. Training such an IDNTM estimation network would typically be straightforward if we had access to a dataset containing both noisy and clean labels. However, since the training datasets of the clients only comprise noisy labels, training the IDNTM estimation network becomes a challenging task, particularly in the FL setting. To tackle this challenge, we introduce a new algorithm called FedBeat, which stands for \underline{\textbf{Fed}}erated Learning with \underline{\textbf{B}}ayesian \underline{\textbf{e}}nsemble-\underline{\textbf{a}}ssisted \underline{\textbf{t}}ransition matrix estimation. FedBeat encompasses three synergistic steps, as illustrates in Figure~\ref{overview}:

\textbf{1. Federated Data Extraction}: This step is used to extract a dataset with high-confidence pseudo-labels to prepare for the training of the IDNTM estimation network. To this end, a weak global classification model is first trained on the noisy labels, which is then used to generate pseudo-labels for the training data samples. To increase the amount of high-confidence pseudo-labels, a Bayesian model ensemble approach is employed to generate pseudo-labels. 

\textbf{2. Federated Transition Matrix Estimation}: The extracted dataset with both high-confidence pseudo-label and noisy labels are used to train the IDNTM estimation network in federated way. When presented with a new data sample, the IDNTM estimation network generates the corresponding IDNTM, where elements represent the transition probabilities from ground-truth labels to noisy labels. 

\textbf{3. Federated Classifier Correction}: With the IDNTM estimation network, we make a correction to the weak global classifier for improvement. In this step, however, the loss function incorporates the output of the IDNTM on the predicted labels, and measures the statistical difference between the actual noisy label and this output.

\section{Problem Formulation}
\subsection{Learning Objective}
We start with the standard $C$-class classification problem in FL, involving $K$ clients and a central server. For each client $k$, it contains the local dataset $\mathcal{D}_{k}= \left\{ \left( \mathbf{x}_{i}, y_{i} \right) \right\}_{i=1}^{N_k}$ where $N_k$ is denoted the size of dataset, i.e. $|\mathcal{D}_{k}|=N_k$, and $y_i \in \{1,2, \dots, C\}$. Each client $k$ has a local model $w^k$ and a model output $f(\mathbf{x}_i,w^k) \in \mathbb{R}^{C\times 1}$, which is the class probability distribution (a.k.a., confidence score vector). The global objective is defined as follows: $\min_x \sum_{k=1}^K {N_k \mathcal{L}_k}/{N} $, where $N = \sum_{k=1}^K N_k$ is the total dataset size. Denote $\boldsymbol{y_i}$ as the one-hot form of $y_i$,  thus $\mathcal{L}_k = {\sum_{i=1}^{N_k}} \mathcal{L} \left( f \left( \mathbf{x}_i,w^k \right), \boldsymbol{y_i} \right) / N_k$ is the local objective function of client $k$, in which $\mathcal{L}(\cdot,\cdot)$ represents the loss function, and $(\mathbf{x}_{i}, y_{i})\in \mathcal{D}_{k}$ is the $i$-th sample of client $k$. 

\subsection{Noise Model}

In the classic FL setting, it is assumed that training samples in each client are labeled with ground-truth. However, this assumption is overly idealistic in real world. In this paper, to capture label noise in real-world data, we consider the IDN model, which assumes that the noise rate is not constant and varies based on each sample's features, rather than varying on each class. Let $D$ be the overall distribution of pairwise random variables $(\mathbf{X}, \mathbf{Y})$, where $\mathbf{X}$ denotes the variables of all training samples stored at clients, and $\mathbf{Y}$ represents the variable of corresponding labels. It can be treated as if each data sample in each client $k$ is independently drawn according to $D$.

Correspondingly, define $\hat{D}$ as the overall distribution of pairwise random variables $(\mathbf{X}, \mathbf{\hat{Y}})$, where $\mathbf{\hat{Y}}$ represents the variable of corresponding noisy labels. The actual dataset held by each client $k$ is denoted as $\mathcal{\hat{D}}_{k}= \left\{ \left( \mathbf{x}_{i},\hat{y}_{i} \right) \right\}_{i=1}^{N_k}$, which depends on the noisy distribution $\hat{D}$. To establish the bridge between the clean distribution $D$ and the noisy distribution $\hat{D}$ and build a statistically consistent classifier, the IDNTM (i.e. $T(\mathbf{x}) = \left(T_{i,j}\left(\mathbf{x}\right)\right)_{i,j = 1}^C \in [0,1]^{C \times C}$) is introduced. Here, we define $T_{i,j} (\mathbf{x})$ as the $i,j$-th element of IDNTM, representing the transition probability of a sample $\mathbf{x}$ with a clean label $i$ transitioning to a noisy label $j$. Thus the noisy class-posterior probability $P(\hat{Y}|\mathbf{X})$ can be inferred by the IDNTM $T(\mathbf{x})$ and the clean class-posterior probability $P(Y|\mathbf{X})$ as follows:
\begin{align}
    P(\hat{Y} = j | X = \mathbf{x}) = \sum_{i=1}^C T_{i,j}(\mathbf{x}) P(Y=i|X=\mathbf{x}).
\end{align}

\section{Proposed Method}
To address FL with instance-dependent noisy labels, we propose FedBeat, which comprises three synergistic steps as follows. 

\subsection{Federated Data Extraction}
As mentioned earlier, we need an IDNTM for each instance transiting from a noisy label distribution to a clean one. Typically, given a subset of instances that have both noisy and clean labels, training an IDNTM estimation network is a relatively straightforward task. However, in an FL context, clients face challenges in accurately labeling each instance, rendering it infeasible to directly build an estimation network for the IDNTM.

Therefore, our first step primarily focuses on training a weak global model from the noisy distribution $\hat{D}$ by incorporating the noisy labels during training to help isolate high-confidence clean data. This initial training lasts for a total of $T_1$ rounds and each client performs $P_1$ local steps during each round. In each round, FedBeat requires the server to disseminate the global model to all clients, which will then fully participate in training and uploading their local models. The server then aggregates the received local models to produce an updated global model by using approaches such as FedAvg \cite{mcmahan2017communication} or FedProx \cite{li2020federated}. We denote the local model uploaded by client $k$ during the final communication round as $w^k_{T_1, P_1}$, and the global model as $\bar{w} = \sum_k N_k w^k_{T_1, P_1} / N$. 

Once the weak global model $\bar{w}$ is obtained, it is returned to the clients to generate peudo-labels on the training data samples. However, rather than returning only $\bar{w}$, we employ a Bayesian model ensemble \cite{chen2020fedbe} approach to boost the classification accuracy, thereby extracting a greater amount of high-confidence data. Particularly, after denoting the average model $\mu = \bar{w}$, the server then calculates the standard deviation $\Sigma$ of the local models:
\begin{align}
\Sigma = \mathrm{diag} \left( \sum\limits_k\frac{N_k}{N} \left( w^k_{T_1, P_1} -\mu \right)^2 \right).
\end{align}
The standard deviation $\Sigma$ is also returned to each client. Here the square operation is element-wise. With both $\bar{w}$ and $\Sigma$, each client $k$ samples $M$ ensemble models ${w^{k, (m)}} \sim \mathcal{N}(\mu, \Sigma),m=1,\dots,M$. It is important to note that the $M$ ensemble models can vary among clients. Such an operation is to derive smoother predictive distributions and minimize the risk of overconfidence. Specifically, each client $k$ computes the average outputs $\Tilde{f}(\mathbf{x}_i)$ (i.e., the average confidence score vector) for each instance $i \in \mathcal{\hat{D}}_k$ using its $M$ ensemble models $w^{k, (m)}$ as follows:
\begin{align}
\Tilde{f}(x_i) = \frac{1}{M}\sum_{m=1}^M f \left(\mathbf{x}_i; w^{k, \left(m\right)}\right),
\end{align}
where $\Tilde{f}(\mathbf{x}_i) \in \mathbb{R}^{C \times 1}$. Let $\Tilde{f}_c(\mathbf{x}_i)$ be the $c$-th elements of $\Tilde{f}(\mathbf{x}_i)$, then the pseudo-label $\Tilde{y}_i$ of $\mathbf{x}_i$ is generated according to $\Tilde{y}_i = \arg\max_c \Tilde{f}_c(\mathbf{x}_i)$, with its associated confidence score being $\Tilde{f}_{\Tilde{y}_i }(\mathbf{x}_i)$. Using the predetermined confidence threshold value $\tau$ (a hyper-parameter), client $k$ filters its local dataset to construct a extracted dataset containing feature inputs (e.g., images), noisy labels, and pseudo-labels:
\begin{align}
\mathcal{\Tilde{D}}_k = \{(\mathbf{x}_i, \hat{y}_i, \Tilde{y}_i); i \in \mathcal{\hat{D}}_k~ \text{and}~\Tilde{f}_{\Tilde{y}_i }(\mathbf{x}_i) \geq \tau\},
\end{align}
where we denote the size of the extracted dataset $\Tilde{D}_k$ as $\Tilde{N}_k$.



With a proper choice of $\tau$, the pseudo-label accuracy is high (see Table \ref{table-threshold}). However, the number of extracted instances can be small compared to the overall dataset size. This limited number presents two challenges: First, training an accurate classification model solely with these instances is impractical. Second, the limited number of filtered instances per client hinders the individual training of an effective IDNTM.


\subsection{Federated Transition Matrix Estimation}
After extracting data with high-confidence pseudo-labels, an IDNTM estimation network $\theta$ is trained using the data extracted on each client. When provided with an input example $\mathbf{x}$, the estimation network $\theta$ generates the IDNTM $\Tilde{T}(\mathbf{x};\theta)$, which contains the probability of transitioning from the clean distribution to the noisy distribution. The training is conducted for $T_2$ rounds, and at each round $t$, a subset of clients is selected to participate in the training. For each selected client $k$, the current version of the estimation network, $\theta_t$, is first received, and it is then set as the local network $\theta_{t,0}^k = \theta_t$. The client then performs $P_2$ steps of local training based on a loss function $\Tilde{\mathcal{L}} (\theta_{t,p}^k)$.
\begin{align}
    \Tilde{\mathcal{L}} (\theta_{t,p}^k)  = - \frac{1}{\Tilde{N}_k}\sum_{i=1}^{\Tilde{N}_k} \sum_{c=1}^C \boldsymbol{\hat{y}_{i}}_{,c} \log \left( \left[ {\boldsymbol{\Tilde{y}_i}^\mathrm{T}} \cdot \Tilde{T} \left(\mathbf{x}_i ; \theta_{t,p}^k \right) \right]^\mathrm{T}_c \right),
\end{align}
where $p \in \{1, \dots, P_2\}$ denotes the local step, the subscript $c$ denotes the $c$-th element of the vector which is usually corresponding to class $c$, $\Tilde{T} (\mathbf{x}_i ; \theta_{t,p}^k)$ is the estimated IDNTM for the example $\mathbf{x}_i$ generated by the local network $ \theta_{t,p}^k$. Here $(\cdot)^T_c$ means we pick the $c$-th element from the $C$-dimensional column vector transposed by the row vector. It is noteworthy that both the noisy label $\boldsymbol{\hat{y}_{i}}$ and the pseudo-label $\boldsymbol{\Tilde{y}_{i}}$ are hard labels, and in the one-hot form of $\hat{y}_{i}$ and $\Tilde{y}_{i}$ respectively. $\Tilde{\mathcal{L}}$ can be viewed as a traditional cross-entropy loss function which treats the noisy label as the ground-truth and $\boldsymbol{\Tilde{y}_i}^\mathrm{T}\Tilde{T} (\mathbf{x}_i ; \theta_{t,p}^k )$ as the predicted output. Such a loss function ensures that the estimated IDNTM, which is generated based on the features of each instance, can successfully map the pseudo-label to the noisy label. Once the selected client completes their local updates, both the updated parameter $\theta_{t,P}^k$ and the size of their extracted dataset $\Tilde{N}_k$ are sent back to the server for further processing. The server then updates the global $\theta_t$ as follows:
\begin{align}
    \theta_{t+1} = \frac{1}{\Tilde{N}}\sum\limits_k\Tilde{N}_k\theta_{t,P_2}^k,
\end{align}
where $\Tilde{N}$ is the sum of the number of all extracted examples. Note that it is crucial to upload $\Tilde{N}_k$ for aggregation on the server instead of using $N_k$ directly as in the federated data extraction step. Though the ratio of aggregation weights among the clients for this process is similar to the previous step in most of IID settings, IID with a high noise rate or non-IID case either causes a huge imbalance in the amount of data extracted. Upon completing $T_2$ rounds, we obtain the final IDNTM estimation network $\theta_{T_2}$, denoted as $\Tilde{\theta}$, which generates the IDNTM that approximates the ground-truth transition matrix.

\subsection{Federated Classifier Correction}
Once the estimation network $\Tilde{\theta}$ has been trained, it can be employed to correct the weak global classification model $\Bar{w}$ derived from the data extraction step. Since $\Bar{w}$ is trained on the noisy dataset $\hat{\mathcal{D}}$, simply continuing to train it in the same way as before would result in overfitting for low noise rates or difficulty in converging for high noise rates. Furthermore, the output distribution would still be based on the noisy dataset.

To obtain a strong classification model, we perform federated classifier correction by conducting $T_3$ communication rounds for global model aggregation, with each round consisting of $P_3$ local steps for fine-tuning the local models similar to the data extraction step. Thus, to minimize the impact of noisy labels, we use the IDNTM estimation network $\Tilde{\theta}$ to map the original predictions to noisy outputs. We define the classifier correction loss function as follows:
\begin{align}
    \Bar{\mathcal{L}} (\Bar{w}_{t,p}^k)  = - \frac{1}{N_k}\sum_{i=1}^{N_k} \sum_{c=1}^C \boldsymbol{\hat{y}_{i}}_{,c} \log \left( \left[ {f\left( \mathbf{x}_i ; \Bar{w}^k_{t,p} \right)^\mathrm{T}} \Tilde{T} \left(\mathbf{x}_i ; \Tilde{\theta} \right) \right]^\mathrm{T}_c \right),
\end{align}
where $\Bar{w}^k_{t,p}$ represents the local model on client $k$ during the $p$-th epoch of round $t$, $f(\mathbf{x}_i ; \Bar{w}^k_{t,p})^\mathrm{T} \Tilde{T}(\mathbf{x}_i ; \Tilde{\theta})$ is the adjusted prediction obtained by mapping the original prediction $f(\mathbf{x}_i ; \Bar{w}^k_{t,p})$ through the IDNTM estimated by the network $\Tilde{\theta}$.

After calculating the loss function for all clients, we aggregate the local models using a global aggregation procedure, similar to the one employed in the federated data extraction step:
\begin{align}
     \Bar{w}_{t+1} = \frac{1}{N}\sum\limits_k N_k \Bar{w}_{t,P_3}^k.
\end{align}
We argue that the federated classifier correction step works by utilizing the IDNTM in reverse. Specifically, the IDNTM produced by $\Tilde{\theta}$ adjusts the clean label distribution to approximate the noisy label distribution. Therefore, we can adjust the output distribution in reverse to approximate a clean label distribution by ensuring that the adjusted output closely matches the noisy label distribution. 

\section{Experiments}
\subsection{Experiment Setup}

\textbf{Baseline Methods.} Given that no previous work has considered IDN in the FL setting, we incorporate several baseline methods to demonstrate the superior performance of our proposed method. We consider three types of baselines. First, we examine some traditional FL methods that learn under the assumption of clean labels (e.g. \textbf{FedAvg} \cite{mcmahan2017communication} and \textbf{FedProx} \cite{li2020federated}). Then, we look at a scenario where each client individually applies the state-of-the-art IDN method developed in centralized learning (\textbf{BLTM} \cite{yang2022estimating}). Lastly, we consider \textbf{FedCorr} \cite{xu2022fedcorr} which is a state-of-the-art FL approach solving CCN.

\textbf{Implementation Details.} Our experiments encompass a range of noise levels (IDN rates: 30\%, and 50\%) across the CIFAR-10 and SVHN datasets. Specific details vary based on the dataset: \textbf{CIFAR-10:}We consider 10 clients, each having 5000 balanced training data samples in the IID setting. In the non-IID scenario, the sample count is scaled down to 60\% of the original. \textbf{SVHN:} In the IID setting, we conisder 10 clients, where each client has 4659 balanced training data samples. In the non-IID setting, the number of samples is reduced to 50\% of the original. We use the Dirichlet distribution ($\alpha_{Dir}$) to simulate the non-IID case among clients.

For the initial step, we employ ResNet-34 \cite{he2016deep} for CIFAR-10 and ResNet-18 \cite{he2016deep} for SVHN as the backbone model with an initial learning rate of 0.01. We warm-up the classification network for a total of 15 rounds, with 10 clients participating in each round, and each client performing 2 local epochs. We then set the number of model ensembles, denoted by $M$, to 10 and the threshold $\tau$ to 0.65. For the second step, which involves training the estimation network, we utilize ResNet-34
for CIFAR-10 and ResNet-18 for SVHN with a 100-dimensional output as the backbone model with an initial learning rate of 0.01. We conduct $T_2$ rounds, set to 85, with 10 clients participating in each round. For the final step, we use the model obtained from the final round of the first step as the backbone model with an initial learning rate of 0.0005. We conduct $T_3$ rounds, set to 50, with 10 clients participating in each round.

\begin{table}[t]
\vspace{-5pt}
  \caption{Test accuracy on CIFAR-10 with different IDN rates}
  \label{table-cifar10}
  \centering
  \resizebox{\linewidth}{!}{
  \begin{tabular}{lllll}
    \toprule
     & \multicolumn{2}{c}{IID} & \multicolumn{2}{c}{non-IID ($\alpha_{Dir} = 1$)} \\
    \cmidrule(r){2-5}
         & IDN-30\% & IDN-50\% & IDN-30\% & IDN-50\% \\
    \midrule
    FedAvg & 73.10 $\pm$ 0.97 & 61.99 $\pm$ 1.82 & 61.41 $\pm$ 3.26 & 47.64 $\pm$ 1.56 \\
    FedProx & 71.97 $\pm$ 1.16 & 58.66 $\pm$ 0.96 & 61.57 $\pm$ 1.65 & 47.74 $\pm$ 0.95 \\
    BLTM-local & 45.75 $\pm$ 0.55 & 36.25 $\pm$ 0.58 & 57.64 $\pm$ 2.01 & 49.13 $\pm$ 1.27 \\
    FedCorr & 65.90 $\pm$ 1.50 & 54.41 $\pm$ 0.89 & 62.23 $\pm$ 2.34 & 50.46 $\pm$ 2.29 \\
    FedBeat(ours) & \textbf{81.58} $\pm$ \textbf{0.24} & \textbf{74.51} $\pm$ \textbf{2.71} & \textbf{72.61} $\pm$ \textbf{0.31} & \textbf{58.44} $\pm$ \textbf{3.53} \\
    \bottomrule
  \end{tabular}
  }
\vspace{-10pt}
\end{table}

\begin{table}
\vspace{-5pt}
  \caption{Test accuracy on SVHN with different IDN rates}
  \label{table-svhn}
  \centering
  \resizebox{\linewidth}{!}{
  \begin{tabular}{lllllll}
    \toprule
     & \multicolumn{2}{c}{IID} & \multicolumn{2}{c}{non-IID ($\alpha_{Dir} = 1$)} \\
    \cmidrule(r){2-5}
         & IDN-30\% & IDN-50\% & IDN-30\% & IDN-50\% \\
    \midrule
    FedAvg & 88.35 $\pm$ 0.91 & 74.24 $\pm$ 0.24 & 83.79 $\pm$ 0.47 & 61.86 $\pm$ 2.98\\
    FedProx & 89.46 $\pm$ 0.23 & 71.81 $\pm$ 2.34 & 84.64 $\pm$ 0.21 & 64.06 $\pm$ 1.13 \\
    BLTM-local & 71.14 $\pm$ 1.39 & 52.26 $\pm$ 0.67 & 70.44 $\pm$ 1.34 & 57.13 $\pm$ 2.78 \\
    FedCorr & 86.18 $\pm$ 3.52 & 70.56 $\pm$ 4.63 & 81.20 $\pm$ 1.61 & 59.37 $\pm$ 3.15 \\
    FedBeat(ours) & \textbf{94.59} $\pm$ \textbf{0.25} & \textbf{87.97} $\pm$ \textbf{2.90} & \textbf{92.59} $\pm$ \textbf{0.40} & \textbf{75.26} $\pm$ \textbf{2.89}\\
    \bottomrule
  \end{tabular} }
  \vspace{-10pt}
\end{table}


\subsection{Comparison with Baseline Methods}
We first evaluate FedBeat against the baseline methods at different noise levels under the IID setting. The results in Tables \ref{table-cifar10} and \ref{table-svhn} from both datasets indicate that our method significantly surpasses the performance of the baselines. Specifically, on CIFAR-10, FedBeat outperforms the baseline methods by at least $8.48\%$ in the IDN-$30\%$ scenario and the performance gap widens to $12.52\%$ in the IDN-$50\%$ scenario. Similar observations can be made from the results on the SVHN dataset. Note that \textbf{BLTM-local} exhibits the worst test accuracy performance under various noise levels. This highlights that directly applying IDN methods developed for centralized learning without suitable cooperation may not yield the desired results in the FL setting. The lackluster performance of \textbf{FedCorr} and \textbf{BLTM-local} emphasizes the effectiveness of our proposed method in FL.

We also perform experiments under the non-IID setting. The results in Tables \ref{table-cifar10} and \ref{table-svhn} show FedBeat, consistently outperforms the baseline methods. Except for \textbf{BLTM-local}, all methods demonstrate a drop in test accuracy on both datasets compared to their performance under IID settings. The stability of \textbf{BLTM-local}'s performance can be attributed to its design, as it does not involve any inter-client cooperation.

Further, we conducted additional non-IID experiments with varying Dirichlet parameters, while keeping the IDN rate fixed at 30\%. These results, presented in Table \ref{table-noniid}, validate that our method consistently performs best under different degrees of non-IID settings. Specifically, FedBeat outperforms all baselines by at least 9.89\% with $\alpha_{Dir} = 0.5$, and by at least 7.49\% with $\alpha_{Dir} = 5$.

\begin{table}[t]
\vspace{-5pt}
  \caption{Test accuracy on SVHN with IDN-30\% varying $\alpha_{Dir}$}
  \label{table-noniid}
  \centering
  \resizebox{\linewidth}{!}{
  \begin{tabular}{llll}
    \toprule
         & $\alpha_{Dir} = 0.5$ & $\alpha_{Dir} = 1$ & $\alpha_{Dir} = 5$ \\
    \midrule
    FedAvg & 81.46 $\pm$ 1.66 & 83.79 $\pm$ 0.47 & 84.98 $\pm$ 0.72 \\
    FedProx & 82.23 $\pm$ 1.42 & 84.64 $\pm$ 0.21 & 84.18 $\pm$ 1.01 \\
    BLTM-local & 73.43 $\pm$ 0.93 & 70.44 $\pm$ 1.34 & 60.08 $\pm$ 0.68 \\
    FedCorr & 76.42 $\pm$ 1.88 & 81.20 $\pm$ 1.61 & 82.61 $\pm$ 2.87 \\
    FedBeat(ours) & \textbf{92.12} $\pm$ \textbf{0.49} & \textbf{92.59} $\pm$ \textbf{0.40} & \textbf{92.47} $\pm$ \textbf{0.31} \\
    \bottomrule
  \end{tabular}
  }
\vspace{-10pt}
\end{table}

\begin{table}[t]
\vspace{-5pt}
  \caption{Impact of model ensemble}
  \label{table-ensemble-w/o}
  \centering
  \resizebox{\linewidth}{!}{
  \begin{tabular}{lll}
    \toprule
         & IDN-30\% & IDN-50\% \\
    \midrule
    w/o ensemble & 92.57\% / 1713 / 90.15 $\pm$ 0.30 & 74.24\% / \textbf{875} / 71.12 $\pm$ 0.84 \\
    w/ ensemble & \textbf{96.01\%} / \textbf{1736} / \textbf{92.59} $\pm$ \textbf{0.40} & \textbf{83.75\%} / 809 / \textbf{75.26} $\pm$ \textbf{2.89} \\
    \bottomrule
  \end{tabular}
  }
\vspace{-10pt}
\end{table}

\begin{table}[!t]
\vspace{-5pt}
  \caption{Impact of threshold}
  \label{table-threshold}
  \centering
  \resizebox{\linewidth}{!}{
  \begin{tabular}{lllll}
    \toprule
     & \multicolumn{2}{c}{IID} & \multicolumn{2}{c}{non-IID ($\alpha_{Dir} = 1$)} \\
    \cmidrule(r){2-5}
         & IDN-30\% & IDN-50\% & IDN-30\% & IDN-50\% \\
    \midrule
    $\tau=0.50$ & 95.49\% / 4112 & 78.81\% / 3917 & 92.28\% / 2008 & 70.05\% / 1876 \\
    $\tau=0.65$ & 97.73\% / 3643 & 90.56\% / 1800 & 96.01\% / 1736 & 83.75\% / 809 \\
    $\tau=0.80$ & 99.21\% / 2362 & 94.77\% / 417 & 97.79\% / 1303 & 85.56\% / 250\\
    \bottomrule
  \end{tabular} }
\vspace{-10pt}
\end{table}


\subsection{Impact of Model Ensemble}
We compare our method with an ablated variant where the first step uses only the average of local models to extract instances with clean pseudo labels. We set the same extraction threshold $\tau$ and ablated variant to isolate the differences between them. As shown in Table \ref{table-ensemble-w/o}, the results (the accuracy of pseudo labeling / the number of extracted data / the final classification accuracy) demonstrate the use of Bayesian model ensembles allows us to filter a similar number of instances with a higher accuracy. This improved accuracy in pseudo-labeling aids the second step in training a more reliable IDNTM estimation network, thereby enhancing the final model's performance.



\subsection{Impact of Threshold}
We further examine the impact of threshold $\tau$. We present the results (the accuracy of pseudo labeling / the number of extracted data) under different $\tau$ values in Table \ref{table-threshold}. If $\tau$ is set too high, the requirement for confidence scores to pass the threshold and be included in the dataset becomes stringent. This could result in too few samples being filtered into the dataset, despite a high correct rate of pseudo-labeling. Conversely, setting a low $\tau$ can result in an excessive number of instances with incorrect pseudo-labels being included in the extracted dataset. Both an excessively small dataset (even with a high accuracy) and a low correct rate (even with a large dataset) can impair the learning performance of the transition network. Therefore, we suggest using a moderate threshold $\tau = 0.65$, which ensures a sufficient number of accurate extracted data.

\section{Conclusion}
In this paper, we studied FL with IDN labels. We introduced a novel algorithm called FedBeat, whose key idea relies on training an IDNTM estimation network to generate the IDNTM that maps ground-truth labels to their corresponding noisy labels. By integrating the IDNTM into the federated training procedure, we aim to enhance the alignment between the ground-truth labels and the model's predicted outputs, ultimately resulting in a global statistically consistent classifier and improved model accuracy. We evaluated the effectiveness of FedBeat through extensive experiments, demonstrating its superiority over state-of-the-art FL algorithms.

\newpage
\bibliographystyle{IEEEbib}
\bibliography{strings,refs}

\end{document}